\documentclass[letterpaper, 10 pt, journal, twoside]{IEEEtran}

\usepackage{xcolor,soul,framed} 

\colorlet{shadecolor}{yellow}
\usepackage{graphicx}
\graphicspath{{../pdf/}{../jpeg/}}

\usepackage{amsmath}
\usepackage{array}
\usepackage[colorlinks,linkcolor=blue]{hyperref}
\usepackage{mdwmath}
\usepackage{mdwtab}
\usepackage{eqparbox}
\usepackage{url}
\usepackage{tabularx} 
\usepackage{float}
\usepackage{parskip}
\usepackage{array}  
\usepackage{multirow}
\usepackage{makecell} 
\usepackage{color}
\usepackage[numbers,sort&compress]{natbib}
 
\usepackage{tabularx}
\usepackage{float}
\usepackage{array}  
\usepackage{booktabs} 
\usepackage{multirow}
\usepackage{makecell}
\usepackage{enumitem}
\setlist[itemize]{noitemsep, topsep=0pt, parsep=0pt, partopsep=0pt,leftmargin=*}

\usepackage{graphicx}
\usepackage{amsmath}
\usepackage{lipsum}
\usepackage{algpseudocode}
\usepackage[ruled, vlined, linesnumbered]{algorithm2e}
\newcommand{\mybluehl}[1]{\textcolor{black}{#1}}

\begin{document}
    \title{No More Mumbles: Enhancing Robot Intelligibility through
Speech Adaptation}
  \author{Qiaoqiao~Ren$^{1}$,
      Yuanbo~Hou$^{2}$, Dick~Botteldooren$^{2}$, and Tony~Belpaeme$^{1}$


\thanks{$^{1}$Qiaoqiao Ren and Tony~Belpaeme are with the Faculty of Engineering and Architecture, IDLab-AIRO, Ghent University -- imec, Technologiepark 126, 9052 Gent, Belgium (e-mail: \{Qiaoqiao.Ren, Tony.Belpaeme\}@UGent.be).}
\thanks{$^{2}$Yuanbo Hou and Dick~Botteldooren are with the Faculty of Engineering and Architecture, WAVES, Ghent University, Belgium (e-mail: \{Yuanbo.Hou, Dick.Botteldooren\}@UGent.be).}

\thanks{The source code, pre-trained model and demos are available at: \href{https://github.com/qiaoqiao2323/robot-speech-intelligibility}{https://github.com/qiaoqiao2323/robot-speech-intelligibility}} 
} 

\markboth{
}{Qiaoqiao \MakeLowercase{\textit{et al.}}: No More Mumbles: Enhancing Robot Intelligibility through
Speech Adaptation}

\maketitle

\begin{abstract}

Spoken language interaction is at the heart of interpersonal communication, and people flexibly adapt their speech to different individuals and environments. It is surprising that robots, and by extension other digital devices, are not equipped to adapt their speech and instead rely on fixed speech parameters, which often hinder comprehension by the user. We conducted a speech comprehension study involving 39 participants who were exposed to different environmental and contextual conditions. During the experiment, the robot articulated words using different vocal parameters, and the participants were tasked with both recognising the spoken words and rating their subjective impression of the robot's speech. The experiment's primary outcome shows that spaces with good acoustic quality positively correlate with intelligibility and user experience. However, increasing the distance between the user and the robot exacerbated the user experience, while distracting background sounds significantly reduced speech recognition accuracy and user satisfaction.  
We next built an adaptive voice for the robot. For this, the robot needs to know how difficult it is for a user to understand spoken language in a particular setting. We present a prediction model that rates how annoying the ambient acoustic environment is and, consequentially, how hard it is to understand someone in this setting.
Then, we develop a convolutional neural network model to adapt the robot's speech parameters to different users and spaces, while taking into account the influence of ambient acoustics on intelligibility. Finally, we present an evaluation with 27 users, demonstrating superior intelligibility and user experience with adaptive voice parameters compared to fixed voice.

\end{abstract}


%
\IEEEpeerreviewmaketitle

\section{Introduction}

\IEEEPARstart{T}{here} is the expectation that social robots will one day be capable of engaging in human-like spoken language interaction with people. 
However, achieving effective spoken interaction with devices such as robots and digital assistants presents a significant challenge, especially in environments with background noise and suboptimal acoustic properties. The parameters that govern a robot's vocal characteristics—such as volume, speech rate, pitch, and emphasis—can impact speech intelligibility and the overall user experience \cite{robinson2022designing}.


Earlier research has underscored the detrimental effects of background noise on speech intelligibility \cite{ishikawa2017effect}. The reverberation time of a space, which strongly correlates with its acoustic quality, has been found to negatively affect speech intelligibility \cite{klatte2010effects}. Additionally, the distance between the listener and the sound source has a crucial impact on speech intelligibility and on the user experience\cite{walters2008human}. 
Earlier studies have identified pitch, pitch range, volume, and speech rate as fundamental vocal characteristics conveying personality \cite{apple1979effects}. For instance, voice pitch has been effectively employed to model a robot's personality and emotion \cite{niculescu2013making}. Speech speed influences speech intelligibility, and Jones \emph{et al.} found that a faster speaking rate lowers comprehension for listeners \cite{jones2007synthesized}. 
Except for the critical role of acoustic factors, individual differences in listeners are also important. Different listeners vary in their ability to understand speech in noisy environments \cite{heinrich2015relationship}. As such, it is crucial to consider the listener's personal information when designing the robot's adaptivity.

\vspace{-0.1cm}
This work advances the field of data-driven adaptive speech, building upon a foundation of prior research that has explored various facets of the challenge, \mybluehl{which has distinct advantages, particularly when addressing complex, variable environments and tailoring individual user needs.} People adapt their speech during spoken conversation. This is known as the Lombard effect, which is an involuntary speech adaptation where speakers naturally modify their speech in noisy environments to improve communication. In response to loud environments, speakers will increase their speech volume, adjust pitch, and enunciate more, thus optimising interaction between interlocutors. However, robots lack this natural adaptability, highlighting the need for advanced systems to mimic the Lombard effect to improve human-robot interaction (HRI). Earlier studies have looked into adapting speech to noisy environments and the distance between the user and the robot \cite{martinson2007improving}, addressing the needs of individuals with hearing impairments \cite{lindley1999adaptation}, or scaling the loudness of the voice dynamically when approaching a user \cite{fischer2021initiating}. Niculescu \emph{et al.} find that pitch adaption of a robot's voice impacted the users' rating of overall interaction quality \cite{niculescu2013making}. However, most robots (and other interactive devices for that matter) do not take the acoustic properties of the environment, the ambient sound, or the user's characteristics into account during the conversation. The large majority of spoken language interactions with robots use fixed voice parameters, often ignoring the dynamic nature of environmental and user factors.

\vspace{-0.05cm}
Our study addresses three research questions (RQs) - RQ1: Is there a relationship between robot speech parameters, user characteristics, acoustic quality of the environment and ambient sound, and the robot's intelligibility as well as user experience? RQ2: Can the robot use information about the environment and the user to adapt its speech parameters? RQ3: Can the robot's speech parameters be dynamically adapted to optimise the user's experience and the robot's intelligibility?

\vspace{-0.05cm}
To address RQ1, we collect data to map how intelligibility suffers under different conditions in Section~\ref{sec:Intelligibility_assessment}. From this, Section~\ref{sec:adaptive_voice} responds to RQ2 by building an automated system to adapt the robot's voice to changing user and environmental factors. Finally, we evaluate our system in a user study in Section~\ref{sec:Evaluation} to address RQ3. Discussion is given in Section~\ref{sec::Discussion}.

\section{Intelligibility assessment}
\label{sec:Intelligibility_assessment}


\subsection{Experimental design}
\label{sec:experiment_design}

\vspace{0.15cm}
\subsubsection{Materials} To reveal how ambient sound, the environment's acoustic quality, the distance between the user and the robot, and the user's hearing can affect both the robot's intelligibility and the user's experience, we first set up a data collection campaign. We used a speech recognition task using a paradigm from \cite{dubno1984effects}. The task relies on open-set recognition of four lists of English consonant-nucleus-consonant (CNC) monosyllabic words (e.g. \emph{hash}, \emph{dodge}, \emph{should}, \ldots) from the Northwestern University Auditory Test 6 (NU-6) \cite{tillman1966expanded}. The words were spoken by an NAO V5 humanoid robot (United Robotics Group, formerly known as Aldebaran Robotics). During the task, a recording of ambient sound (ranging from soft chatter to the sound of power tools) was played through a speaker placed near the robot. The voice parameters of the robot (i.e., volume, speech rate, pitch, and emphasis) were drawn using a uniform distribution from predetermined parameter ranges, thus sampling the voice parameter space. 


\begin{figure}[h]
    \centering{\includegraphics[width=0.8 \columnwidth]{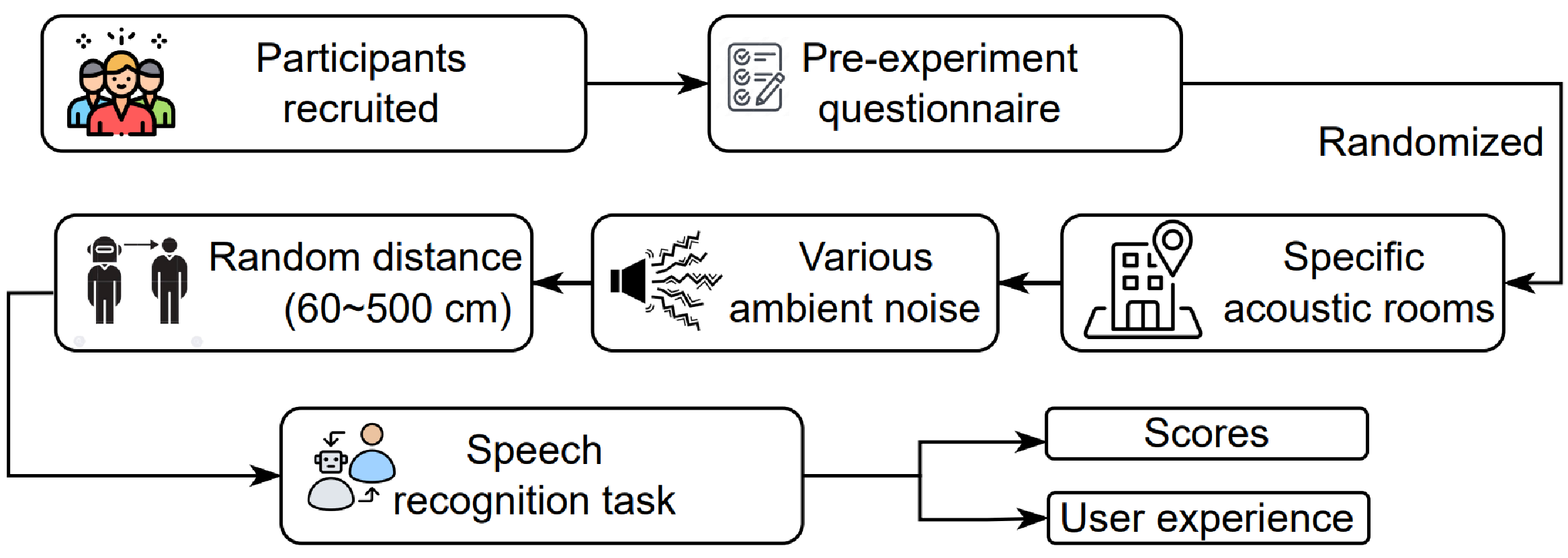}}
    \caption{Experimental procedure.}
    \label{ex_design}
\end{figure}

\vspace{0.2cm}
\subsubsection{Procedure} 
\begin{figure}[h]
	\setlength{\belowcaptionskip}{0cm} 
    \centering{\includegraphics[width=\columnwidth]{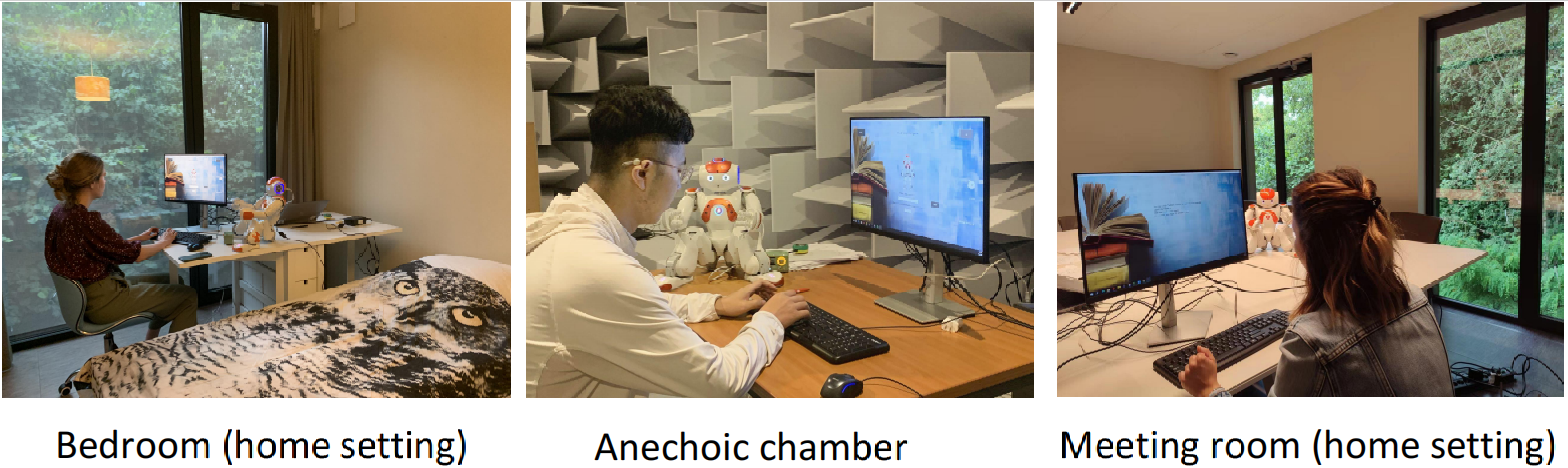}}
    \caption{Illustration of different environments in which data was collected.}
    \label{robot.eps}
\end{figure}

Participants were recruited via a social media campaign in the local area, excluding those with self-reported reading-related learning difficulties, to ensure data quality. The data collection and study adhered to the ethics procedures of the \emph{Universiteit Gent}. Participants gave informed consent and were told they could withdraw at any time without providing a reason. The experiment process is shown in the Fig.~\ref{ex_design}. A pre-test questionnaire gathered personal information, including gender and age, their self-evaluated hearing difficulties in noisy environments, and their self-reported Common European Framework of Reference for Languages (CEFR) levels as their English levels. 39 participants (17 identified as female, 20 as male, 2 as other; $28.0\pm3.7$ years old) took part in the data collection exercise. Each participant was randomly assigned to one of six different rooms, each having its own acoustic quality. The participants were exposed to three different ambient sounds drawn from a total of 11; complete silence was used as a baseline condition. After, the participants were placed at a random interaction distance to the robot, ranging from 60 to 500 \textit{cm}. Each participant is exposed to 50 words per round; During the task, the robot spoke a word picked randomly from the NU-6 list and the participant was asked to type the word using a keyboard, along with providing feedback on the user experience. Fig.~\ref{robot.eps} shows examples of different settings in which data was collected.

Next, we build a model to examine how different factors influence the intelligibility of the robot's speech and the user experience. As independent variables, we have two participant characteristics, four robot voice parameters, and three environmental factors. As dependent variables, we collected similarity scores and user experience from the participants per word. Therefore, each word is a tuple consisting of the independent variables and dependent variables. We collected 5442 tuples in total.

\vspace{-0.2cm}
\subsection{Factors influencing intelligibility}
\label{sec:independent}

\vspace{0.2cm}
\subsubsection{Participant characteristics}

We collect data on participants' hearing loss and English language level. Individuals with normal hearing already experience difficulties in comprehending speech in noisy environments, and this is exacerbated when people experience hearing loss. We therefore ask participants to self-report hearing loss. 
Moreover, listening in adverse conditions presents more challenges for non-native listeners in a foreign language context \cite{lecumberri2010non}. Thus, we also record their English proficiency level. Specifically, participants rated their difficulty hearing in noisy environments, such as restaurants or social gatherings, using a Likert scale (1: No, never, to 5: Yes, always), \mybluehl{which has been used in previous research \cite{mikkola2015self}} and rated their English proficiency using the CEFR scale, which rates English proficiency from A1 (1: beginner) to C2 (6: fluent). This yields distributions for hearing difficulty ($2.2 \pm 1$) and English proficiency level ($4.8 \pm 0.8$).

\vspace{0.3cm}
\subsubsection{Robot voice parameters}
The second data category relates to four specific parameters associated with the robot's vocal characteristics: speed, pitch, emphasis, and volume. These features play a crucial role in determining the clarity and intelligibility of the robot's speech. We study their impact on participants' experiences and speech intelligibility.

\vspace{-0.1cm}
The words are produced using the built-in US English speech engine of the NAO robot, which is supplied by Nuance. The designers of NAO state that the default voice is the voice of a machine but with a fluidity that makes it very comfortable to listen to \cite{kali2021walking}, and the default voice has been widely used in research studies \cite{tozadore2017wizard}. We modulate four key parameters in the voice engine: volume, speed, pitch, and emphasis (which is implemented by double voice synthesis and controlled by the \textit{doubleVoiceLevel} parameter, which adds a second voice over the first voice). The volume ranges between [0, 2], and the pitch ranges between [0.5, 2], with a default of 1. Speed is set as a multiple of the default rate of 100\%, ranging between [0.4, 4], and governs the pace at which the speech is articulated. Lastly, the \textit{doubleVoiceLevel}, ranging between [0, 4], grants control over the gain of a second voice; 0 disables the effect. In the word recognition task, all the voices are played with NAO's loudspeakers set to 100\%. The default voice settings for the voice parameters—volume, speed, and pitch are all set at 1, with the double voice effect deactivated.





\vspace{0.3cm}
\subsubsection{Environmental factors}
 
The third category of data comprises three environmental factors. The reverberation time (T30), the participants' annoyance rating (AR) of the ambient noise, and the distance of the individuals to the robot. 

\begin{figure}[H]
    \centering
    \includegraphics[width=0.8\columnwidth]{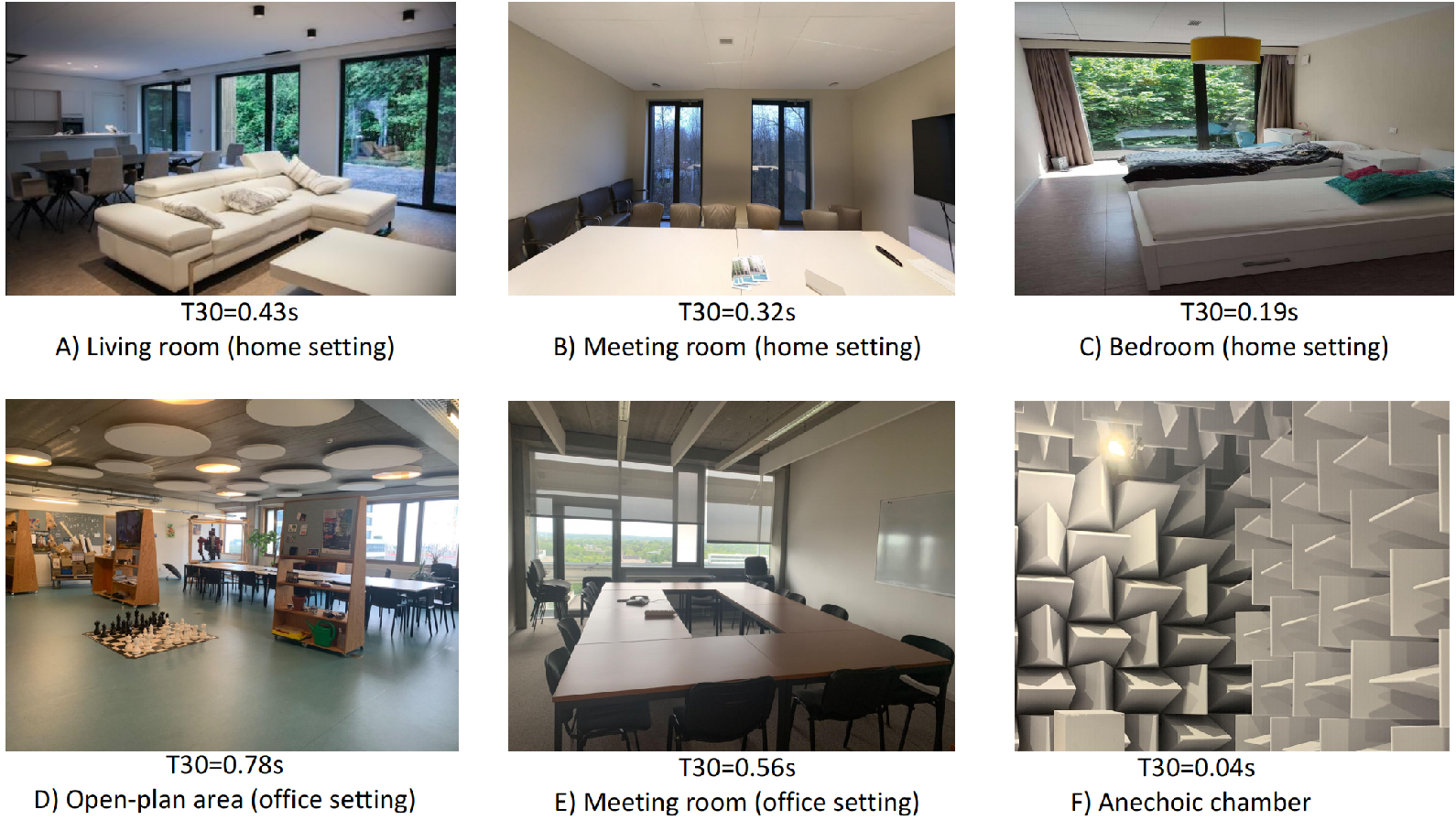}
    \caption{Reverberation time (T30) of different rooms.}
    \label{rooms.eps}
\end{figure}

\vspace{-0.1cm}
Reverberation time is a key metric for assessing a room's acoustic quality. The T30 is a standard metric and reports the time it takes for the sound pressure level to decrease by 60 decibels \cite{paini2006reverberation}, as detailed in the ISO3382 standard. To measure the T30, an impulse noise (a popping balloon) was used, and the reverberation time was measured using a Svantek SVAN 959 Sound and Vibration Analyser. During the measurement, the impulse source and microphone were 1.0m above floor level, and we measured at three positions for each room (the middle of the room and two diagonal corners). The T30 measures were averaged over a frequency range from 250Hz to 2kHz based on the three measurements per room 
\cite{dobreva2011influence}. The measured results are given in Fig.~\ref{rooms.eps}. For the data collection and evaluation, we used different spaces, each with varying reverberation times (in seconds(s)). Data was collected in an office setting, which includes an open-plan area (T30 $=0.78$s), meeting room (T30 $=0.56$s), and a home setting (the imec Homelab) consisting of a living room (T30 $=0.43$s), bedroom (T30 $=0.19$s) and meeting room (T30 $=0.32$s), and an anechoic chamber (T30 $=0.04$s) used for echo-free sound assessment. Those environments offer a stable setting without disturbance from uncontrollable variables like crowds.

\vspace{-0.1cm}
Next to the different rooms, we also make use of different ambient sounds. The subjective perception of these sounds-- and specifically the extent to which these sounds are perceived as annoying, which is correlated with the volume of the sounds-- is expected to influence the participant's performance on tasks \cite{vastfjall2002influences}. The ambient noise played in the experiment is from the DeLTA dataset, which includes 2980 samples of 15-second binaural audio recordings with 24 classes of sound sources and human-annotated annoyance ratings~\cite{mitchell_andrew_2022_7158057}. The annoyance rating ranges from 1 (not annoying at all) to 10 (very annoying). All the ambient sounds are played at the same speaker volume, guaranteeing a consistent playback level.
For the robot to quantify the subjective experience of ambient sounds, it needs a model that maps a sound recording into an annoyance rating. Important here is that the model not only looks at the amplitude of the ambient sound but also takes a holistic view of how annoying the sound is. 

\vspace{-0.1cm}
Finally, the distance participants keep from the robot will impact the robot's intelligibility. During data collection, we sampled distances ranging from 60 to 500 \textit{cm}.

\subsection{Intelligibility metrics}
\label{sec:dependent}

The intelligibility metrics encompass both user experience and speech intelligibility.

\vspace{0.2cm}
\subsubsection{User experience}
Participants' user experience related to the robot's voice is also collected. This is rated from 1 (not pleasant at all) to 10 (very pleasant). These measures help gauge the affective and subjective aspects of the interaction.

\vspace{0.1cm}
\subsubsection{Speech intelligibility}
Participants were tasked with typing the words spoken by the robot. We evaluated the robot's speech intelligibility by measuring the phonetic similarity score between the spoken words and the words typed by the participants. This compensates for homophones, such as \emph{flour} and \emph{flower}, and gives us a scalar metric for how poorly the word was understood. We used a phonetic embedding method to calculate the phonetic similarity of test and input words from participants, as in \cite{sharma2021phonetic}. This phonetic embedding represents each phonetic component as a vector in a continuous vector space, thus encoding phonetic information, considering phoneme alignment in the words and capturing intricate phonetic characteristics and enhancing similarity assessment.

\vspace{-0.2cm}
\subsection{Generalised linear mixed-effect models}
Based on the independent variables collected in Section~\ref{sec:independent} and dependent variables from Section~\ref{sec:dependent}, we build a generalised linear mixed-effect model (GLMM) to study the contributions of various factors to spoken language intelligibility and to the user experience, including fixed effects, interact effects and random effects. We use the method by \cite{nelder1972generalized}, which can handle non-normal distributed response variables. \mybluehl{A Variance Inflation Factors analysis revealed an absence of multicollinearity among the independent variables - all factors were below the threshold ($\leq 2$). In addition, the correlations between the annoyance rating and acoustic quality, as well as the user information, are all smaller than 0.1.}

The gamma distribution is a continuous probability distribution to model data that is skewed, continuous and non-negative \cite{moran2007new}, such as the residuals of the phonetic scores and user experience that we observed. To choose the most accurate model, we compare the prediction error using the Akaike Information Criterion (AIC) of the reverse link function and the log link function. We chose the log link function, as it obtained the lowest AIC.

\vspace{-0.1cm}
The generalised model examines the relationship between predictor variables and both the intelligibility and user experience, as represented by Eq~\ref{SP_Gamma_GLMM} and Eq~\ref{UX_Gamma_GLMM}. To adequately account for the non-independence of observations stemming from repeated measurements across subjects and the inherent variability in word difficulty, we have integrated subject ID and word ID as random effects in our models to control for intra-subject variability and variable difficulty levels of the words tested.

 \vspace{-0.4cm}
\begin{align}
\label{SP_Gamma_GLMM}
&\text{SP}_{ij} \sim \text{Gamma}(\mu_{ij}, \phi) \nonumber \\
&\log({\mu_{ij}+ \mu}) = \beta_{0} \cdot \text{v}_{ij} + \beta_{1} \cdot \text{p}_{ij} + \beta_{2} \cdot \text{e}_{ij} + \beta_{3} \cdot \text{s}_{ij}  \nonumber \\
&\quad + \beta_{4} \cdot \text{A}_{ij} + \beta_{5} \cdot \text{T}_{ij} + \beta_{6} \cdot \text{D}_{ij} + \beta_{7} \cdot \text{E}_{ij}  + \beta_{8} \cdot \text{H}_{ij} \nonumber \\
&\quad+ \beta_{9} \cdot (\text{v} \cdot \text{T}) + 1|pn_{ij}+ 1|w_{ij} + \epsilon_{1}  \nonumber \\
&\epsilon_{1} \sim \text{Normal}(0, \sigma^2)
\end{align}

\vspace{-6mm}

\begin{align}
\label{UX_Gamma_GLMM}
&\text{UX}_{ij} \sim \text{Gamma}
(\lambda_{ij}, \phi) \nonumber  \\
&\log(\lambda_{ij}) = \lambda_{0} \cdot \text{v}_{ij} + \lambda_{1} \cdot \text{p}_{ij} + \lambda_{2} \cdot \text{e}_{ij} + \lambda_{3} \cdot \text{s}_{ij}  \nonumber \\
&\quad + \lambda_{4} \cdot \text{A}_{ij} + \lambda_{5} \cdot \text{T}_{ij} + \lambda_{6} \cdot \text{D}_{ij} + \lambda_{7} \cdot \text{E}_{ij} + \lambda_{8} \cdot \text{H}_{ij} \nonumber \\
&\quad + \lambda_{9} \cdot (\text{v} \cdot \text{T}) + 1|pn_{ij} + 1|w_{ij} + \epsilon_{2}  \nonumber \\
&\epsilon_{2} \sim \text{Normal}(0, \sigma^2)
\end{align}

Eq~\ref{SP_Gamma_GLMM} and Eq~\ref{UX_Gamma_GLMM} use the coefficients \(\beta_{0}\) to \(\beta_{9}\) and  \(\lambda_{0}\) to \(\lambda_{9}\) respectively to represent the estimated impact of predictor variables on the response variables (speech intelligibility, user experience). They show how each predictor is expected to log-linearly impact the response, and these effects are fixed across the dataset, not subject to random variation within the model's framework. Specifically, the coefficients \(\beta_{0}\) to \(\beta_{3}\ , \lambda_{0}\) to \(\lambda_{3}\) correspond to the voice attributes (volume, pitch, emphasis, and speed) of the robot on speech intelligibility and user experience, respectively, while \(\beta_{4}\) to \(\beta_{6}\ , (\lambda_{4}\) to \(\lambda_{6}\)) represent factors such as room characteristics (AL, T30) and distance to the robot. English proficiency of participants and their auditory challenges in noisy environments are signified by \(\beta_{7}\) to \(\beta_{8}\) and \(\lambda_{7}\) to \(\lambda_{8}\) for user's intelligibility and user experience, respectively. The coefficients \(\beta_{9}\) pertain to the interaction of volume and the room's acoustic quality and the influence on both speech intelligibility and user experience. Similarly, \(\lambda_{9}\) represents the interactions in the context of user experience. Simultaneously, the unaccounted-for residuals that evade fixed and random effects are encapsulated by the residual component \(\epsilon\). The subscript \(ij\) indicates the \(j\)th observation within the \(i\)th group. The response variable \(SP_{ij}\),  \(UX_{ij}\) embodies speech intelligibility and user experience in this context, respectively, and $\mu$ is $0.0001$. 


\vspace{-0.1cm}
We used the SIMR package in R for post-hoc power analysis, based on the 1000 Monte Carlo simulations \cite{green2016simr}. As the log link function was used in GLMM, we did the exponential transform of the coefficients ( \(\beta_{0}\) to \(\beta_{9}\ , \lambda_{0}\) to \(\lambda_{9}\) ) to get the estimate of the fixed effect for speech intelligibility and user experience.

\vspace{-1mm}
\subsection{Intelligibility model results}

\vspace{2mm}
\subsubsection{Speech intelligibility model analysis}

The negative estimate of $-0.75$ associated with \textit{annoyance rating} underscores the adverse impact of heightened annoyance on speech intelligibility ($p < 0.001$), the post-hoc power is 98.60\%. Additionally, the \textit{acoustic quality of the space (T30)} has a significant negative estimate of  $-0.77$  with speech intelligibility ($p < 0.01$), the post-hoc power is 66.90\%; this finding signifies that good acoustic conditions of a space are conducive to comprehension of the robot's speech. Additionally, the \textit{speed} has a significant negative estimate of $-0.21$  ($p < 0.001$) of speech intelligibility, which obtained 100\% post-hoc power. Finally, \textit{pitch} has a negative estimate of $-0.71$ ($p<0.001$), which implies that the high pitch leads to decreased intelligibility with 91.70\% post-hoc power. We observed some weak estimates without statistical significance  ($p>0.05$). For instance, there is a negative correlation between the \textit{distance}, \textit{emphasis} as well as \textit{hearing difficulty} and speech intelligibility. In terms of \textit{English proficiency level}, which serves as an indicator of language skills, our analysis revealed a marginally positive coefficient.

\vspace{0mm}
We explored the interplay between the \textit{T30} of spaces and the \textit{volume} of the robot's voice and its impact on speech intelligibility, the results suggest a significant positive estimate of $1.44$ interaction effect between room characteristics and the volume of the robot's voice ($p < 0.001$) with 76.20\% power, indicates that the relationship between volume and speech intelligibility is influenced by the quality of room acoustics, and the optimal volume for improve the robot speech intelligibility depending quality of the spatial acoustic, which might explain why the post-hoc power for \textit{T30} is relatively lower. This may indicate that the effect size of it is trivially small compared with the interaction effect. The random effect attributed to subjects and words ID accounts for smaller than 1\% of the overall variance for speech intelligibility. This indicates that the influence of the random effect, including the subject and word ID, on the total variability of speech intelligibility is modest.


\vspace{3mm}
\subsubsection{User experience model analysis}


The results showed that the predictor \textit{annoyance rating} exhibited a negative estimate of $-0.74$ ($p < 0.001$) and the post-hoc power is 83.00\%, signifying that heightened annoyance ratings due to ambient sounds correspond to a decrease in user experience. Likewise, the predictor \textit{pitch} displayed a significant negative coefficient of $-0.89$ ($p < 0.05$) with 65.6\% post-hoc power, indicating that elevated pitch levels are associated with lower user experience scores. This suggests that an increased pitch might have an adverse impact on user experience during such interactions. The predictor \textit{distance} also yielded a significant negative coefficient of $-0.74$ ($p < 0.01$) with 95.7\% post-hoc power, implying that greater distances from the sound source are linked to lower user experience scores. Furthermore, the predictor \textit{T30} of space revealed a negative coefficient of $-0.67$ with statistical significance ($p < 0.01$) and the post-hoc power is 81.30\%. This implies that the quality of the room environment significantly influences user experience, with a decrease in room quality corresponding to a decrease in user experience. Surprisingly, the \textit{English proficiency level of users exhibited a significant negative} estimate of $-0.54$ with user experience ($p < 0.01$) with 75.20\% post-hoc power. This implies that individuals with higher language proficiency experienced the robot's speech as less pleasant. This might be attributed to their heightened expectations regarding word intelligibility. In addition, the predictor \textit{speed} ($p < 0.001$) displayed a significant negative coefficient of $-0.56$ with 99.6\% post-hoc power, indicating that faster speech rates result in a poorer user experience.

\vspace{-1mm}
Interestingly, the predictor \textit{hearing difficulty} in noisy environments did not predict user experience ($p > 0.05$). This suggests that participants who reported higher hearing difficulty experienced a significantly better user experience listening to robot speech. In contrast, the predictor \textit{emphasis} showed a negative estimate with user experience in this analysis ($p > 0.05$), although it was not statistically significant. In the context of user experience, the direct effect of volume on user experience showed a significant negative relationship (fixed estimate of -0.92, $p < 0.01$), suggesting that higher volume levels are generally perceived negatively. However, this finding comes with an important caveat as the statistical power for this effect was notably low (18.8\%), indicating potential limitations in the robustness of this result. Our finding indicated the significant interaction between room acoustics and the robot's volume (coefficient 1.32, $p < 0.001$), with a robust post-hoc power of 83.3\%. The positive interaction coefficient indicates that the optimal volume for enhancing user experience shifts depending on the acoustic quality of the room. This interaction effect potentially explains the low power observed in the direct effect of \textit{volume}, which shows that the robot voice should be adapted to the environment to get better user experience and intelligibility. The random effect attributed to individual subjects and word ID accounts for less than 5\% of the overall user experience variance. This suggests the user experience is similar for all subjects and all words.

\section{Building an adaptive robot voice}
\label{sec:adaptive_voice}

One of the key takeaways from our analysis is the importance of adapting the robot's voice to the environment and user. To create an adaptive robot voice, we required a model to assess ambient sound annoyance. Then, we constructed an adaptive speech model that leverages the annoyance rating, and together with other parameters-- the distance to the robot, the T30 reverberation of the room, the user's English proficiency, and hearing difficulties-- adjusts the robot's speech parameters.

\vspace{-2mm}

\subsection{Ambient sounds' annoyance rating prediction}

\begin{figure}[h] 
\centering
	\setlength{\abovecaptionskip}{0cm}  
	\setlength{\belowcaptionskip}{0cm}   
 \rotatebox{90}{\includegraphics[width = 0.15 \textwidth]{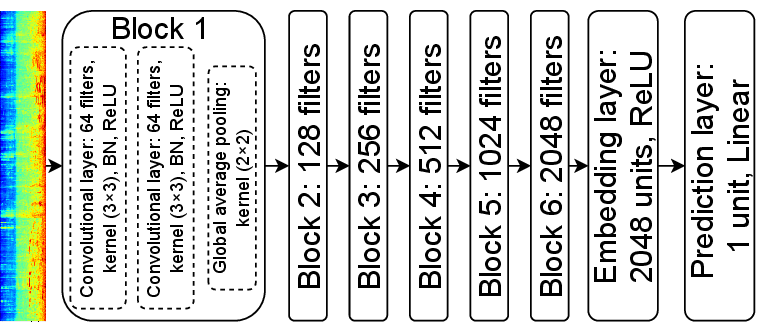}}
	\caption{The CNN-based ARP model for overall evaluation of ambient sounds.}
	\label{annoyance_model}
\end{figure}

As ambient sound is known to strongly impact the intelligibility of speech and user experience, we require an automated method to rate the annoyance of ambient sounds. This not only takes into account the amplitude of the sound, but also takes into account the frequency spectrum of ambient sound to predict the annoyance rating (AR). This is a scalar value used as an overall evaluation metric to reflect the impact of ambient sounds on participants' ability to understand speech. Fig.~\ref{annoyance_model} shows the convolutional neural network (CNN)--based annoyance rating prediction (ARP) model we use for this.

\label{ssec:figure-f}

Inspired by the outstanding performance of the convolution-based sound source classification model \cite{kong2020panns}, the ARP model uses 6 convolutional blocks of the same structure but with a sequentially increasing number of filters. 
Each convolutional block consists of two convolutional layers equipped with ReLU activation functions and a global average pooling layer \cite{relu}. To stabilise and speed up the neural network training, batch normalisation (BN) \cite{batchnormal} is introduced. 
After the convolutional blocks, an embedding layer consisting of fully connected layers with a dimension of 2048 is used to transform ambient sound representations extracted by convolutional blocks into high-level embeddings suitable, for the final prediction layer of the whole environmental sound.

\vspace{2mm}
\subsubsection{Ambient sound ARP model training} 

To train the ARP model to successfully infer the impact of environmental sounds on participants, i.e., the annoyance rating for participants, based on acoustic features containing amplitude, frequency, and category information of ambient sounds, we trained the model on the DeLTA real-life polyphonic audio dataset  \cite{mitchell_andrew_2022_7158057}. The DeLTA contains 2980 samples of 15-second binaural audio recordings with 24 classes of sound sources from European cities for noise annoyance detection. Participants rated 15-second binaural recordings of urban environments by providing an annoyance rating on a scale of 1 to 10. 

\vspace{-1mm}
As shown in Fig.~\ref{annoyance_model}, given the prediction output from the final prediction layer is $\hat{y_{ar}}$, and its corresponding label is ${y_{ar}} \in [1, 10]$, the mean squared error (MSE) is used as the loss function for the ambient sound ARP model, $Loss = MSE(\hat{y_{ar}}, y_{ar})$.
To comprehensively consider polyphonic audio information such as the amplitude and frequency of ambient sound sources, the acoustic feature \textit{log Mel} that performs well in sound-related tasks is adopted. \textit{Log Mel}-filter 64-bank spectrogram is extracted by the Short-Time Fourier Transform with a Hamming window length of 46\textit{ms} and a window overlap of $1/3$ \cite{rgasc}, which means that the duration of each frame is 0.031\textit{s}, and the acoustic features can capture the acoustic information of sounds with a time resolution of 0.031\textit{s}. 
In training, dropout and normalisation are used to prevent over-fitting of the model. 
The Adam optimiser \cite{adam}, with an initial learning rate of 0.005, minimises the loss function, with a batch size of 64. The model is trained for 100 epochs. 
In inference, using a Tesla V100-SXM2-32GB GPU and Intel(R) Xeon(R) CPU E5-2698 v4@2.20GHz as an example, the response time of the model from inputting acoustic features to outputting prediction results is 1.87\textit{ms}, which means the model can process input sounds in real-time.

\subsubsection{Ambient sound ARP model performance}
We used 2200 clips from the DeLTA dataset to train the model, retaining 245 clips for validation and 445 for testing. Table \ref{tab:models_result} shows the performance of the proposed CNN-based ARP model on the test set, and also shows the performance of several other typical neural network models. 
Among them, the deep neural network (DNN) consists of four fully connected (FC) layers and the ARP layer. Each FC layer's number of units is 64, 128, 256, and 512, respectively. The final FC layer's output is flattened and input to the ARP layer. 
The Simple CNN consists of 4 convolutional layers, each equipped with (3 × 3) kernels, and the ARP layer. Each convolutional layer's filter numbers are 64, 128, 256, and 512, respectively. The final convolutional layer's output is flattened and input to the ARP layer. 
The CNN-Transformer consists of 3 convolutional layers with (3 × 3) kernels, a Transformer encoder \cite{Transformer}, and the ARP layer.
In Table \ref{tab:models_result}, the results of the simple CNN are slightly better than those of the DNN, illustrating the effectiveness of the convolutional layer in extracting information such as the amplitude and frequency of ambient sounds from the input acoustic features. The CNN-Transformer, equipped with a Transformer encoder suited for temporal modelling, performs better than a simple CNN. Finally, the proposed CNN model achieved more competitive results with lower MSE and MAE. This shows that the proposed CNN-based ARP model can effectively predict the impact of ambient sound based on its comprehensive information on the participants. 



\begin{table}[H]  
	\setlength{\abovecaptionskip}{0.2cm}   
	\setlength{\belowcaptionskip}{-0.2cm}  
	\renewcommand\tabcolsep{1pt} 
	\centering
	\caption{The ARP performance on the test set of the DeLTA dataset.}
	\begin{tabular}{
	p{1.cm}<{\centering}| 
	p{1.cm}<{\centering}|
 p{1.7cm}<{\centering}|
 p{2.3cm}<{\centering}|
 p{1.8cm}<{\centering}
	} 
 \specialrule{0.1em}{1.5pt}{1.5pt}
	Model	&  DNN &  Simple CNN & CNN-Transformer & Proposed CNN \\
\specialrule{0.1em}{1.5pt}{1.5pt}
 MSE	 &  1.73 & 1.67 & 1.45 &  \textbf{1.10} \\
\specialrule{0.1em}{1.5pt}{1.5pt}
 MAE	 &  1.01 & 1.00 & 0.97 &  \textbf{0.83}\\
	 \hline 
	\end{tabular}
	\label{tab:models_result}
\end{table}

\vspace{-6mm}
\subsection{Proposed adaptive speech model}
\label{sec:adaptive_model}

Our findings emphasise that it is crucial to tailor the robot's voice to the specific environment to achieve both optimal speech intelligibility and a positive interaction experience for different users. We trained a neural network to predict optimal robotic voices based on user and environment characteristics. The network is trained only on data for which participant speech recognition was completely accurate and user satisfaction was greater than 5. This excludes data of negative user experiences or where the robot's speech was poorly understood. We call this model the Environment-to-Voice model (ETV). It takes environmental factors as input, including annoyance ratings of ambient sound (reported by the ARP model), the distance to the robot, the English proficiency of the user, the user's difficulty hearing in noisy environments, and the room's acoustic quality. The model's output consists of robot voice parameters, which consist of volume, speed, pitch, and emphasis.
The network has 5 input nodes, 2 hidden layers with a  ReLU activation function (16 and 32 neurons, respectively) and a 4-neuron output layer with a linear activation function.

\vspace{2mm}
\textit{Model training}: Data augmentation is used to expand the dataset and introduce noise. We use the Adam optimiser \cite{adam}, with a default learning rate of 0.0001. The model trains for 200 epochs, achieving a minimal MSE of 0.15.

\begin{figure}[h]
    \centering
    \includegraphics[width=1\columnwidth]{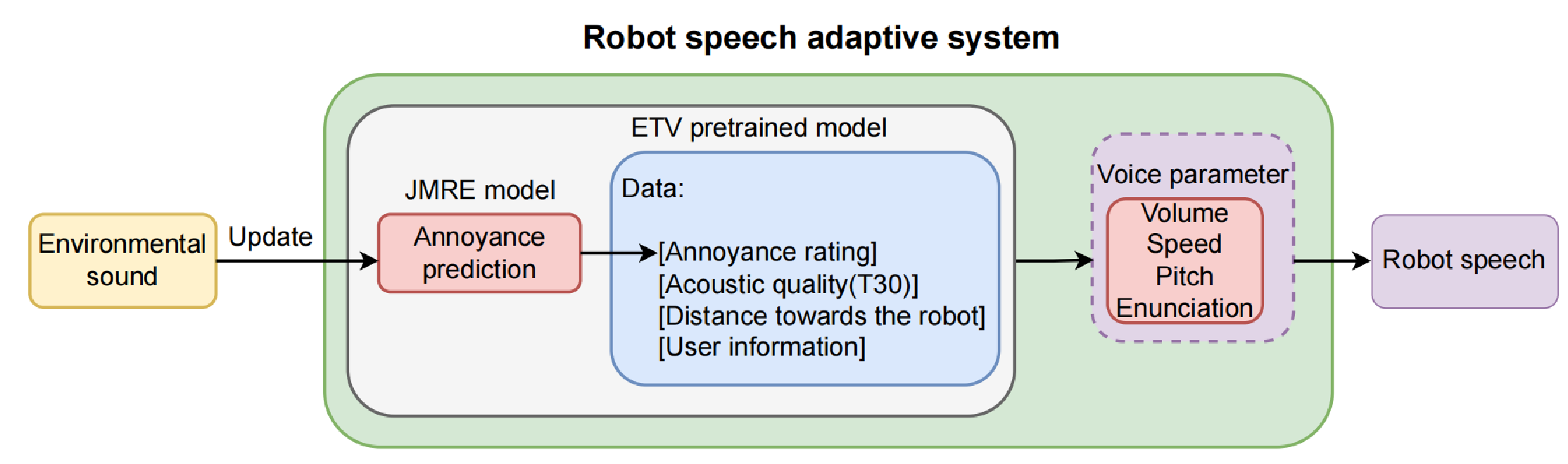}
    \caption{Robot speech adaptive system.}
    \label{adaptive}
\end{figure}

We evaluated the adaptive robot voice using the method reported in section \ref{sec:experiment_design}. The ARP model is used to predict environmental annoyance ratings via the robot. Notably, due to the use of a lightweight model, the computation, despite running on the CPU, only requires 2 seconds. Subsequently, the predictions are communicated to a pre-trained ETV model. This predictive data, together with user-provided information on hearing difficulties and English proficiency and assessments of acoustic quality, serves as input for the ETV model, which then generates adaptive voice parameters. The robot speech adaptive system is visually depicted in Fig.~\ref{adaptive}.

\vspace{-0mm}
\section{Evaluation}
\label{sec:Evaluation}

\vspace{2mm}
\subsection{Evaluation experiment design}

To assess the generalisation of our proposed ETV model with unseen data, we conducted an evaluation experiment involving 27 participants (14 identified as female, 13 as male; $26.6\pm2.4$ years old, Hearing difficulty: $2.4\pm0.6$, English proficiency level: $4.8\pm0.92$). The experiment followed a within-subject design, using two conditions - \textit{(1) Adaptive Speech}: The robot's speech parameters were dynamically adjusted based on the environmental and spatial settings. \textit{(2) Fixed Speech}: The robot uses its default voice parameters.

Each participant experienced both conditions in a balanced random order. The evaluation experiment took place in two rooms with T30 values of $0.04$ and $0.56$, respectively. Participants were asked to type the words spoken by the robot within one of these two rooms. Within each condition, there were two sessions featuring distinct ambient sounds, which are high annoyance rating (AR) ($AR\geq 5$) and low AR ($AR < 5$). In each session, the robot spoke 15 words randomly drawn from the NU-6 list.

\begin{figure}[b]
    \centering
 \includegraphics[width=0.7\columnwidth]{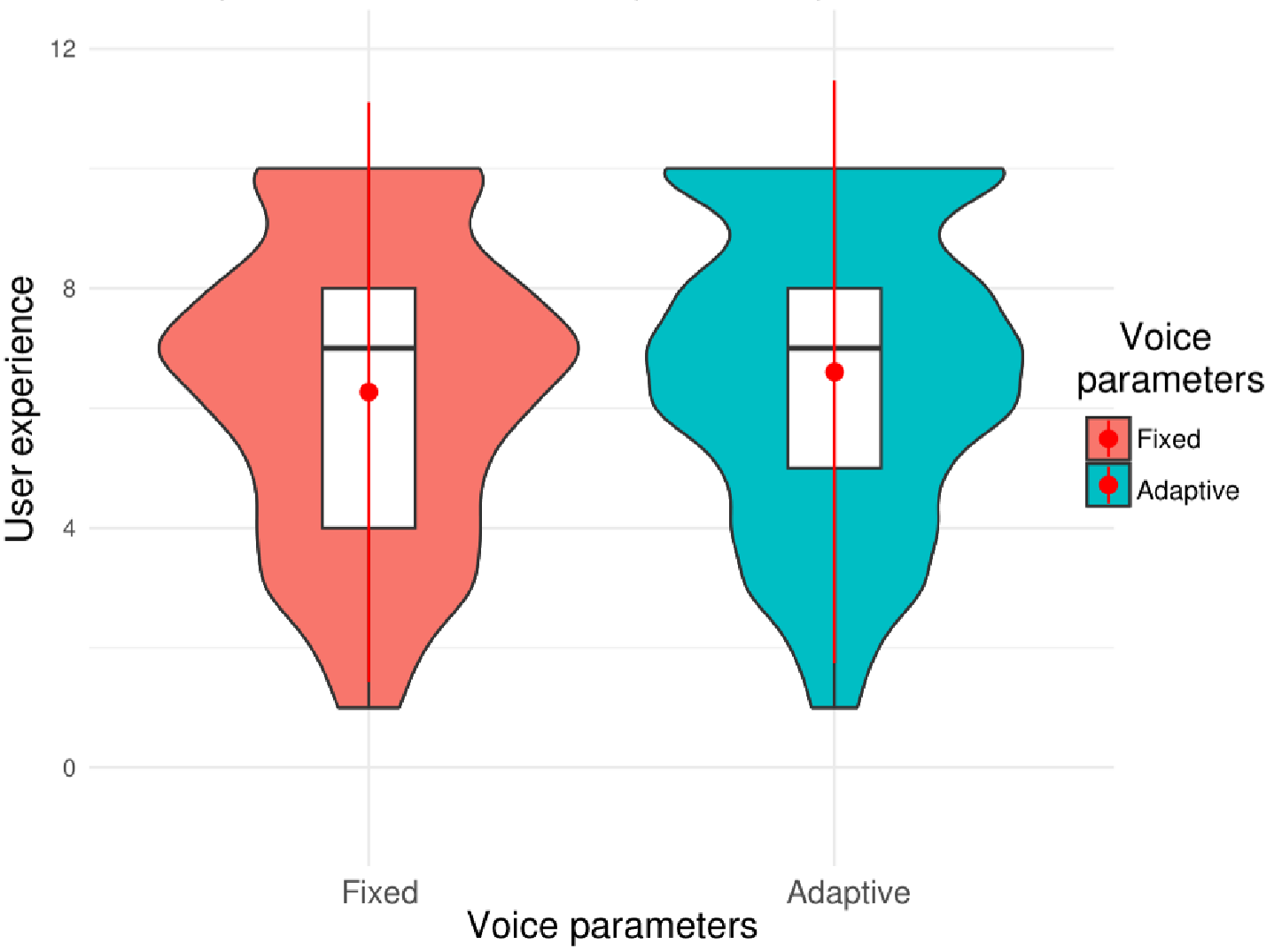}
    \caption{User experience ratings for default robot voice (left) and the adaptive voice (right).}
    \label{UX_E.eps}
\end{figure}

\vspace{-2mm}
\subsection{Fixed robot voice versus adaptive voice}

\begin{figure}[b]

    \centering
    \includegraphics[width=0.7\columnwidth]
    {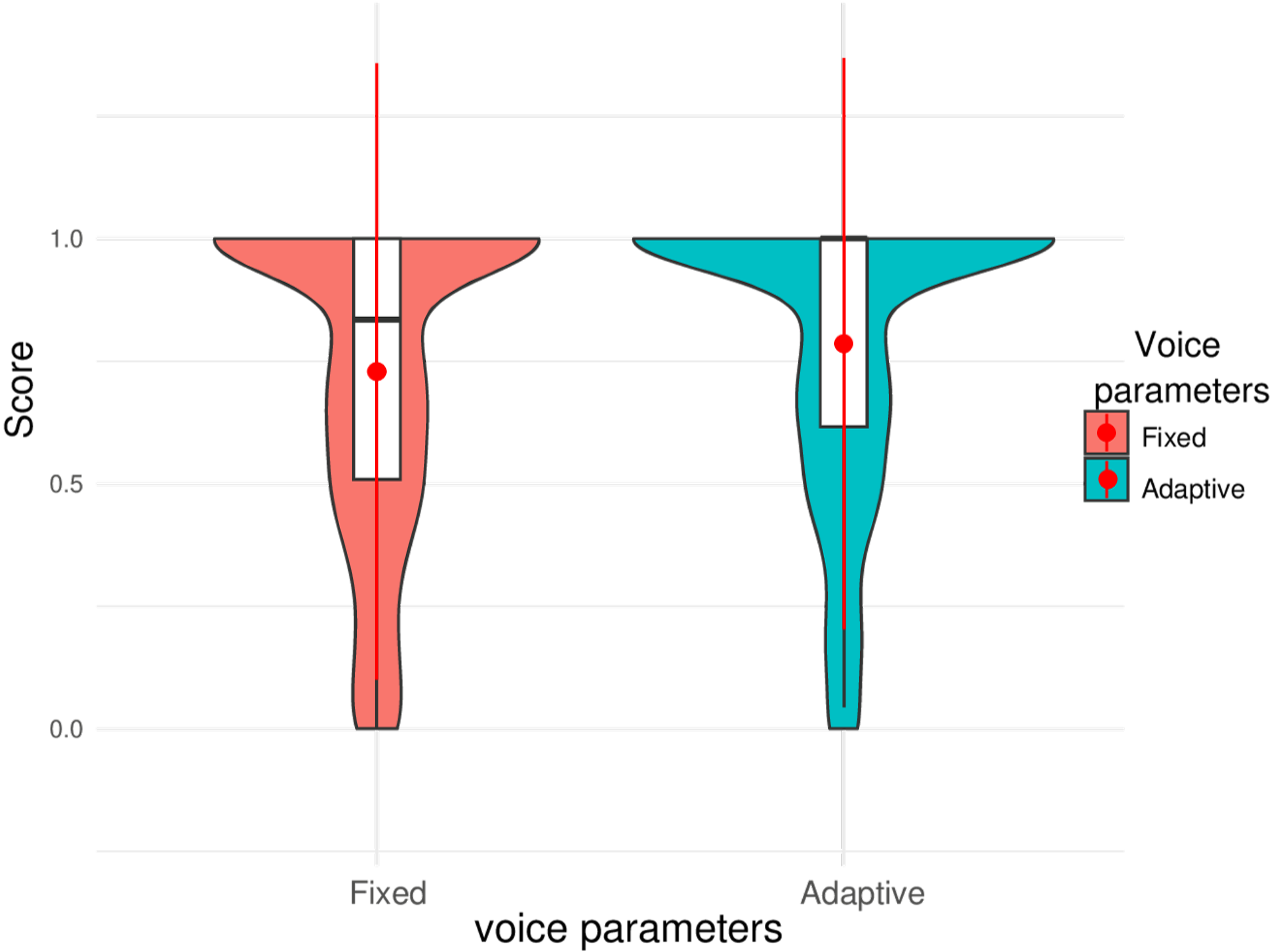}
    \caption{Intelligibility of the robot, expressed as phonetic similarity scores, when it used the default voice (left) or its adaptive voice (right).}
    \label{PS_E.eps}
\end{figure}

In our within-subject design study, the phonetic similarity scores and user experience data did not follow a normal distribution; hence we used the non-parametric Wilcoxon Signed-Rank Test; this analysis yielded an effect size (r), demonstrating that the users' intelligibility of the robot's speech is significantly better ($Z= -3.79, p < 0.001$) for the adaptive robot voice than for the fixed robot voice. The adaptive robot voice also leads to a significantly better user experience ($Z = -2.99, p = 0.003$). This can be seen in Fig.~\ref{PS_E.eps}, Fig.~\ref{UX_E.eps} and Tab~\ref{tab_evaluation}; The post-hoc power is calculated using G*Power \cite{gpower} with 27 participants. The phonetic similarity scores post-hoc power is $97.5\%$ with $\alpha = 0.05$ and $r = 0.73$, and user experience post-hoc power is $88.8\%$ with $\alpha = 0.05$ and $r = 0.58$. The horizontal line that splits the box in two is the median, which in Fig.~\ref{PS_E.eps} coincides with the top line; the mean is indicated by the red dot. The top and bottom boundaries of the box indicate the 25th and 75th percentiles, respectively.

\begin{table}[H] 
    \caption{Means and Standard Deviations for Phonetic Similarity with fixed and adaptive robot voice parameters}
    \begin{tabular} 
    {
	p{1.2cm}<{\centering}
	p{2.3cm}<{\centering}|
 p{1.5cm}<{\centering}|
 p{2cm}<{\centering}
	} 

            \specialrule{0.1em}{1.5pt}{1.5pt} 
        AR & Voice parameters  & Scores & User experience\\
        (1-10) & adaptive Vs fixed  & Mean $\pm$ Std & Mean $\pm$ Std \\
               \specialrule{0.1em}{1.5pt}{1.5pt}
        \multirow{2}{*}{All} & fixed & 0.73 $\pm$ 0.31 & 6.27 $\pm$ 2.42 \\
          & adaptive & 0.79 $\pm$ 0.29 & 6.60 $\pm$ 2.43 \\
        \specialrule{0.1em}{1.5pt}{1.5pt}
        \multirow{2}{*}{Low AR} & fixed & 0.78 $\pm$ 0.28 & 6.60 $\pm$ 2.36 \\
          &adaptive & 0.81 $\pm$ 0.28 & 6.86 $\pm$ 2.32 \\  
           \specialrule{0.1em}{1.5pt}{1.5pt}
 
        \multirow{2}{*}{High AR} & fixed & 0.67 $\pm$ 0.33 & 5.93 $\pm$ 2.43 \\
        &adaptive & 0.76 $\pm$ 0.30 & 6.34 $\pm$ 2.52 \\
       \specialrule{0.1em}{1.5pt}{1.5pt}
         \label{tab_evaluation}
    \end{tabular}
\end{table}

\vspace{-9mm}
\subsection{Fixed robot voice under different annoyance ratings environment sound}

As all the participants experienced the two different kinds of ambient sounds, i.e. low annoyance and high annoyance sounds, the Wilcoxon Signed-Rank Test was used to evaluate the differences between fixed robot voice and adaptive voice under high AR and Low AR, respectively. The results can be seen in the Tab.~\ref{tab_evaluation}. For high AR, we observed that the users rated significantly higher pleasantness for adaptive voice than the fixed default voice setting ($Z= -4.01, p < 0.001$) and significantly better scores are observed for the adaptive voice than the default voice ($Z= -3.82, p = 0.009$). However, there is no significant difference between the adaptive voice and the default voice on the scores under the low AR. Interestingly, the users rate the adaptive voice to be significantly more pleasant than the default voice ($Z= -2.33, p = 0.043$). This indicates that in the case of low ambient noise, although the adaptive sound does not improve the intelligibility, it does improve the user experience.

\vspace{-2mm}
\section{Discussion}
\label{sec::Discussion}

\subsection{Recommendations}

As expected, \mybluehl{there is a relationship between robot speech parameters, user characteristics, environment factors, and the robot’s intelligibility as well as user experience.} Specifically, we find that annoying sounds in the environment make it harder to understand the robot's speech, which in turn hurts the user experience. This emphasises the need for adaptive robot speech in noisy settings. In addition, the room's acoustic quality influences the intelligibility and the user experience, highlighting the need to take the room into account. Moreover, increased distance will worsen the user experience. Therefore, it might be interesting to further explore the proper distance for spoken conversation between a robot and its user. The results show that, as expected, the default and therefore nonadaptive robot voice will lead to subpar speech intelligibility and user experience when there is ambient sound in the environment. In brief, a robot voice that does not adapt to the user and the environment is a missed opportunity.

Regarding the factors related to the user, unsurprisingly, English proficiency matters. People understand the English words spoken by the robot better when they are more proficient in English. Surprisingly, they report having a lower interaction experience, which we hypothesise might be explained by their having stricter pronunciation expectations for the robot. Similarly, the user's hearing difficulties will exacerbate their understanding of the robot. These outcomes strongly suggest that user characteristics should be taken into consideration when setting the robot's voice and speech parameters.

As for robot speech parameters, a higher volume will help intelligibility, but worsen the user experience. Unsurprising, as shouting makes you heard, but it is not pleasant. Slow speech will improve both the user experience and the intelligibility. Surprisingly, robots with higher pitch values lead to poorer satisfaction and lower intelligibility. One possible explanation is that users may find sharp voices unpleasant or may not appreciate the emotional implications conveyed by higher pitch levels. For instance, earlier research noted that menacing voices often exhibit an increase in both volume and pitch. Additionally, increased pitch levels are commonly associated with expressions of anger \cite{murray1993toward}. 
Surprisingly, better emphasis ---implemented through adding a double voice --- had a small negative influence on both intelligibility and user experience. This effect could be attributed to the anomaly of emphasising a single word without a context. 

Designing adaptive, user-centric robot systems that consider the individual user and environmental conditions is essential for effective communication in HRI. This requires the robot to be informed about the environment and the user, ideally in an automated way. The distance between the user and the robot was now manually entered into the model, but could, of course, be extracted from an RGBD camera. T30 reverberation estimates could also be approximated by the robot through speech \cite{ratnam2003blind}  during the conversation or by playing an impulse sound when entering a new environment and analysing the reverberation. The ARP model predicts the impact of ambient noise. Other factors, such as the user's language proficiency or potential hearing difficulties are harder to automatically extract but could be collected during the making of a user profile. \mybluehl{In addition, from the evaluation experiment, we can conclude that the adaptive robot voice optimises the user’s experience and the robot’s intelligibility than its default voice.}

\subsection{Limitations}

This study has certain limitations. First, the data we collected suffers from a class imbalance across predictors, as people with hearing difficulties or very low English proficiency levels are underrepresented. This imbalance can introduce bias into the model's outcomes. Secondly, the participants may not be representative of the broader population, thus limiting the generalizability of the findings. Future studies could draw from a more diverse user population.
The experimental design, involving both within-subject and between-subject factors, introduces an additional complexity as not all group conditions for environmental factors are fully covered within the study. \mybluehl{In addition, even though the self-reported hearing difficulties or English levels has been used in previous research, it might still be influenced by individuals's perception.}

The speech recognition task relies on recognising single words, which is a difficult task. Embedding words in a sentence would improve intelligibility due to the availability of context. Our data was collected under controlled conditions, and a range of factors typical of real-world HRI scenarios, such as the presence of other speakers or contextual cues, were not taken into account. Our findings may not fully extend to more complex scenarios. However, it serves as a foundation for optimizing robots for more complicated scenarios.

One participant noted that it might be advantageous to inquire about participants' listening and English \emph{writing} abilities specifically, as the task exclusively involves listening and writing rather than an assessment of overall English proficiency; in addition, the different participant's accent or background could influence their perception of words, especially for non-native speakers, who might struggle with understanding short phrases and single words. Finally, the position of the robot, whether on the left or right side of the participants, may influence their intelligibility of the robot's speech. 

\subsection{Conclusion}

Given our results, it is difficult to understand why virtually all current robot voices do not adapt to the environmental context and the diversity in users. Our research sheds light on how adaptive voices can have a positive impact on both the user experience and speech intelligibility. Simply put, shouting louder to be better understood is not the solution. Instead, a judicious balance is needed between making the robot understood and not irritating the user.  
In light of these, we presented the ETV model that seeks to enhance robot speech intelligibility by predicting and setting user-adapted and contextually suitable robot voice parameters, tailored to users, and to spatial and environmental conditions. Important to note is that the use of a data-driven model alleviates the need to use heuristics to change the robot's voice.

\bibliographystyle{IEEEtran}
\bibliography{RAL_voice}
\end{document}